\documentclass[11pt]{article}

\usepackage[final]{acl}

\usepackage{times}
\usepackage{latexsym}

\usepackage[T1]{fontenc}

\usepackage[utf8]{inputenc}

\usepackage{microtype}

\usepackage{inconsolata}

\usepackage{graphicx}

\usepackage{url}            
\usepackage{booktabs}       
\usepackage{nicefrac}       
\usepackage{microtype}      
\usepackage{xcolor}         
\usepackage{amsfonts}       
\usepackage{mathtools}      
\usepackage{bbm}            
\usepackage{amssymb}        
\usepackage{multirow}       
\usepackage{siunitx}        
\usepackage{enumitem}       
\usepackage{fontawesome5}
\usepackage{amsmath}
\usepackage{caption}        
\usepackage{float}
\usepackage{adjustbox}
\usepackage{wrapfig}
\usepackage{inconsolata}
\usepackage[most]{tcolorbox}
\usepackage{xspace}
\usepackage{array}
\usepackage{makecell}
\usepackage{tikz}
\usepackage{xparse}
\usepackage{listings}
\usepackage{algorithm}
\usepackage{algpseudocode}
\usepackage{framed}
\usepackage{adjustbox}
\usepackage{import}
\usepackage{xifthen}
\usepackage{pdfpages}
\usepackage{transparent}
\usetikzlibrary{arrows.meta, positioning, shapes.geometric, calc, patterns, shapes.callouts}

\input{macros.tex}

\title{\methodname: Principled Information Seeking\\via Evidence Retrieval and Strategic Questioning}

\author{%
  Maksym Taranukhin$^{1,2}$~~~Shuyue Stella Li$^4$~~~Evangelos Milios$^5$\\\textbf{Geoff Pleiss}$^{1, 2,3}$~~~\textbf{Yulia Tsvetkov}$^4$~~~\textbf{Vered Shwartz}$^{1,2,3}$\\\\
  $^1$ University of British Columbia\qquad
  $^2$ Vector Institute\qquad
  $^3$ CIFAR AI Chair\\
  $^4$ University of Washington\qquad
  $^5$ Dalhousie University\\
  \texttt{\small \{maksymt, vshwartz\}@cs.ubc.ca, \{stelli, yuliats\}@cs.washington.edu,}\\[-0.2em]
  \texttt{\small geoff.pleiss@stat.ubc.ca, eem@dal.ca}
}

\begin{document}
\maketitle
\begin{abstract}
LLMs are increasingly deployed in high-stakes domains such as medical triage and legal assistance, often as document-grounded QA systems in which a user provides a description, relevant sources are retrieved, and an LLM generates a prediction. In practice, initial user queries are often underspecified, and a single retrieval pass is insufficient for reliable decision-making, leading to incorrect and overly confident answers. While follow-up questioning can elicit missing information, existing methods typically depend on implicit, unstructured confidence signals from the LLM, making it difficult to determine what remains unknown, what information matters most, and when to stop asking questions. We propose \methodname, a framework that gathers missing information from two complementary sources: retrieved domain documents and targeted follow-up questions to the user. \methodname models uncertainty using Dempster-Shafer belief assignments over a structured evidential network, enabling principled fusion of incomplete and potentially contradictory evidence from both sources without prematurely collapsing to a definitive answer. Across legal and medical tasks, \methodname outperforms strong baselines while requiring fewer turns. By grounding uncertainty in formal evidential theory rather than heuristic LLM signals, \methodname moves towards trustworthy, interpretable decision support in domains where reliability is critical.
\end{abstract}

\section{Introduction}
\label{sec:intro}
LLMs are increasingly deployed in high-stakes domains such as healthcare and law, where mistakes can have serious consequences \cite{shui-etal-2023-comprehensive,hager_evaluation_2024,lawSurvey, kelsallRapidEvidenceReview2025}. 
Errors are particularly likely when user queries are underspecified \citep{min-etal-2020-ambigqa}. Therefore, LLM-based conversational agents must effectively recognize and address information gaps necessary for robust reasoning \cite{li2024mediq, mayne-etal-2025-llms}. However, current LLM training paradigms incentivize confident responses to all questions, including those that are incomplete or ambiguous, rather than encouraging the admission of uncertainty or the generation of clarifying questions \cite{kapoor2024large,zhang-etal-2024-r,wang-etal-2025-learning}.

\begin{figure}[t]
    \centering
    \includegraphics[width=\linewidth]{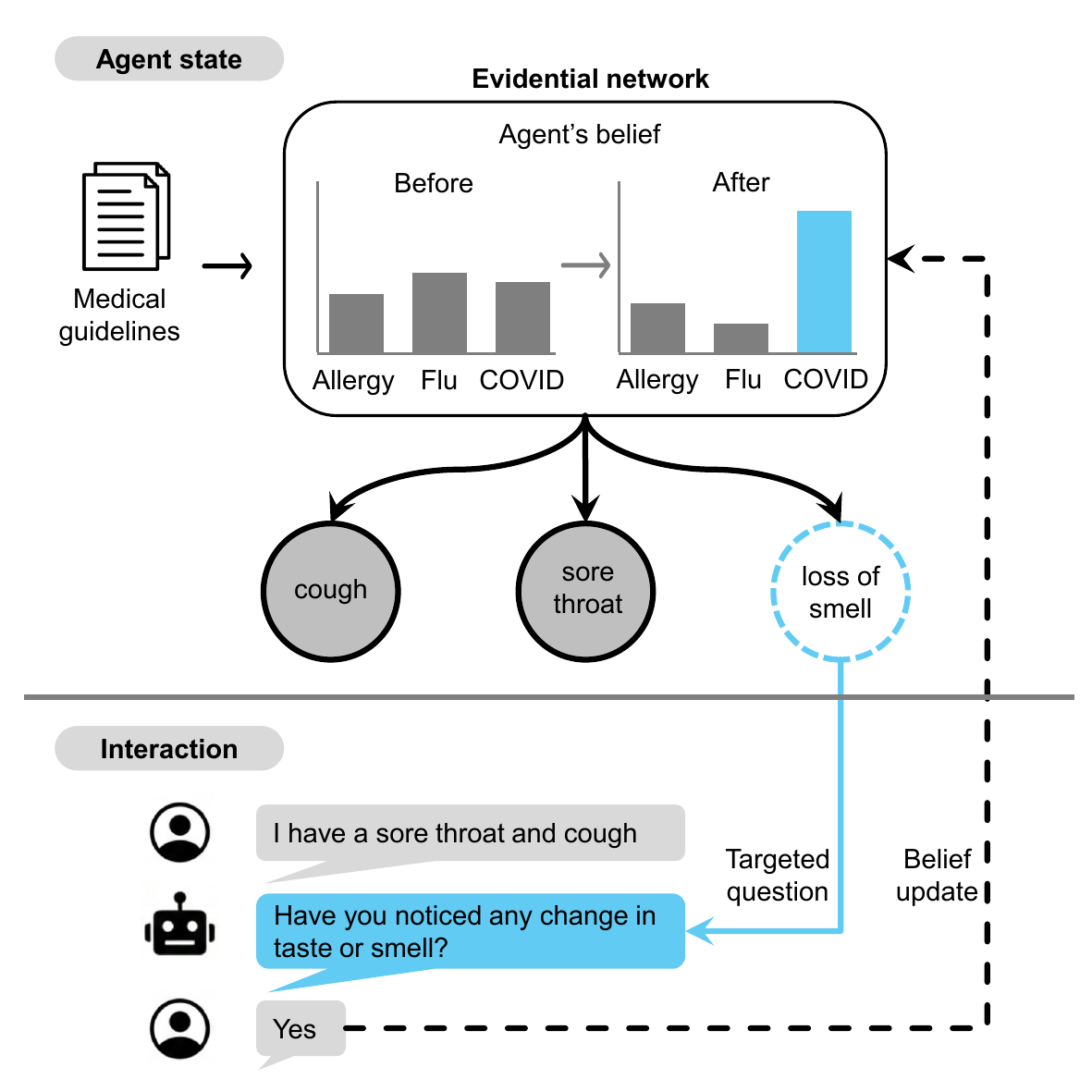}
    \caption{The overview of the \methodname method. The agent uses user input and relevant documents to construct an evidential network representing its current belief in the true hypothesis. By identifying areas of uncertainty within the network, the agent generates targeted follow-up questions to refine its belief systematically.}
    \label{fig:overview}
\end{figure}

A natural way to address this shortcoming is through information-seeking dialogues \cite{li2024mediq, lee-etal-2024-redefining}, where models can ask clarification questions \cite{rao-daume-iii-2019-answer, aliannejadi2019asking,majumder-etal-2021-ask}. Prior work encouraged information-seeking behaviour in LLMs via fine-tuning \citep{li2025alfa,wu2025webdancer,sun-etal-2025-simpledeepsearcher,huang-etal-2025-teaching} or prompting \cite{pan2023kwaiagents, hu2024uncertainty}, but lacked a principled framework for identifying uncertainty to guide the questioning strategy. Such systems may ask unnecessary questions or pose them in suboptimal order, leading to inefficient dialogues that either miss critical information or overwhelm the user with irrelevant queries. 

In contrast, existing uncertainty-guided methods estimate confidence directly from the LLM's output logits or verbalized probabilities \cite{feng2025bird}. These metrics, however, are often uncalibrated and prone to overconfidence, frequently failing to accurately reflect the model's true uncertainty about context completeness \cite{tian-etal-2023-just, NEURIPS2024_1bdcb065}.

Furthermore, effective information-seeking requires access to reliable domain knowledge. However, most existing methods depend on the LLM's parametric knowledge, which may be incomplete, outdated, or derived from unreliable sources. Consequently, these systems often fail to acquire necessary information, thereby constraining informed decision-making in complex, specialized domains.

To address these limitations, this work introduces \methodname, a framework that applies Dempster-Shafer evidential reasoning to conversational information gathering. This approach enables principled uncertainty modeling over evidence derived from both retrieved domain documents and user responses. As illustrated in Figure~\ref{fig:overview}, when presented with an incomplete user query ``I have a sore throat and cough'', \methodname retrieves relevant documents from medical guidelines, constructs an evidential network over competing hypotheses (`Allergy', `Flu', and `COVID'). It then analyzes the network to identify the most informative evidence gap and generates a targeted follow-up question about the loss of smell. The user's response refines the agent's initial belief, steering it toward the true hypothesis, COVID.

\methodname differs from previous approaches in two principal ways. First, it models uncertainty using Dempster–Shafer belief assignments rather than classical point probabilities, thereby allowing explicit representation of ambiguity and ignorance without prematurely converging on a definitive answer. Second, it grounds interactions in authoritative documents rather than relying on the LLM's parametric knowledge, which may be incomplete, outdated, or unreliable. As a result, the agent asks more relevant questions and accurately determines when it has gathered enough data to make the final decision.

We evaluate \methodname in two high-stakes domains. In medicine, we use MedQA \cite{jin2021disease}, a multiple-choice benchmark for clinical knowledge and diagnostic reasoning. In law, we use BarExamQA \cite{zheng2025}, which assesses legal reasoning and professional judgment across a broad range of bar-exam topics. In both settings, a client provides an incomplete medical or legal problem, and the agent must identify the correct answer by asking follow-up questions. \methodname delivers substantial gains on both benchmarks while asking fewer questions and is therefore more efficient. By endowing LLM agents with principled uncertainty representation and document-grounded reasoning, \methodname moves towards trustworthy, domain and task-agnostic conversational agents that can be deployed broadly to support laypersons and experts in decision-making processes.

\section{Method}
\label{sec:method}
We formalize \textit{Strategic Information Acquisition} (SIA) as an active information-gathering process over a latent, finite hypothesis space $\mathcal{H}$, contextualized by an auxiliary corpus $\mathcal{D} = \{d_1, \dots, d_n\}$. An agent aims to identify the true hypothesis $h^* \in \mathcal{H}$ through interaction with a user. At interaction step $t$, the agent maintains an internal belief state $b_t$ over $\mathcal{H}$ which is based on both \static{} information from $\mathcal{D}$ and \dynamic{} information from the user responses. The agent generates a question $q_t$ to elicit a user response $a_t$, which is used for updating the belief state $b_{t+1}$. The process is repeated until sufficient confidence in a given hypothesis is achieved. 

\begin{figure*}[htbp]
    \centering
    \includegraphics[width=0.48\textwidth]{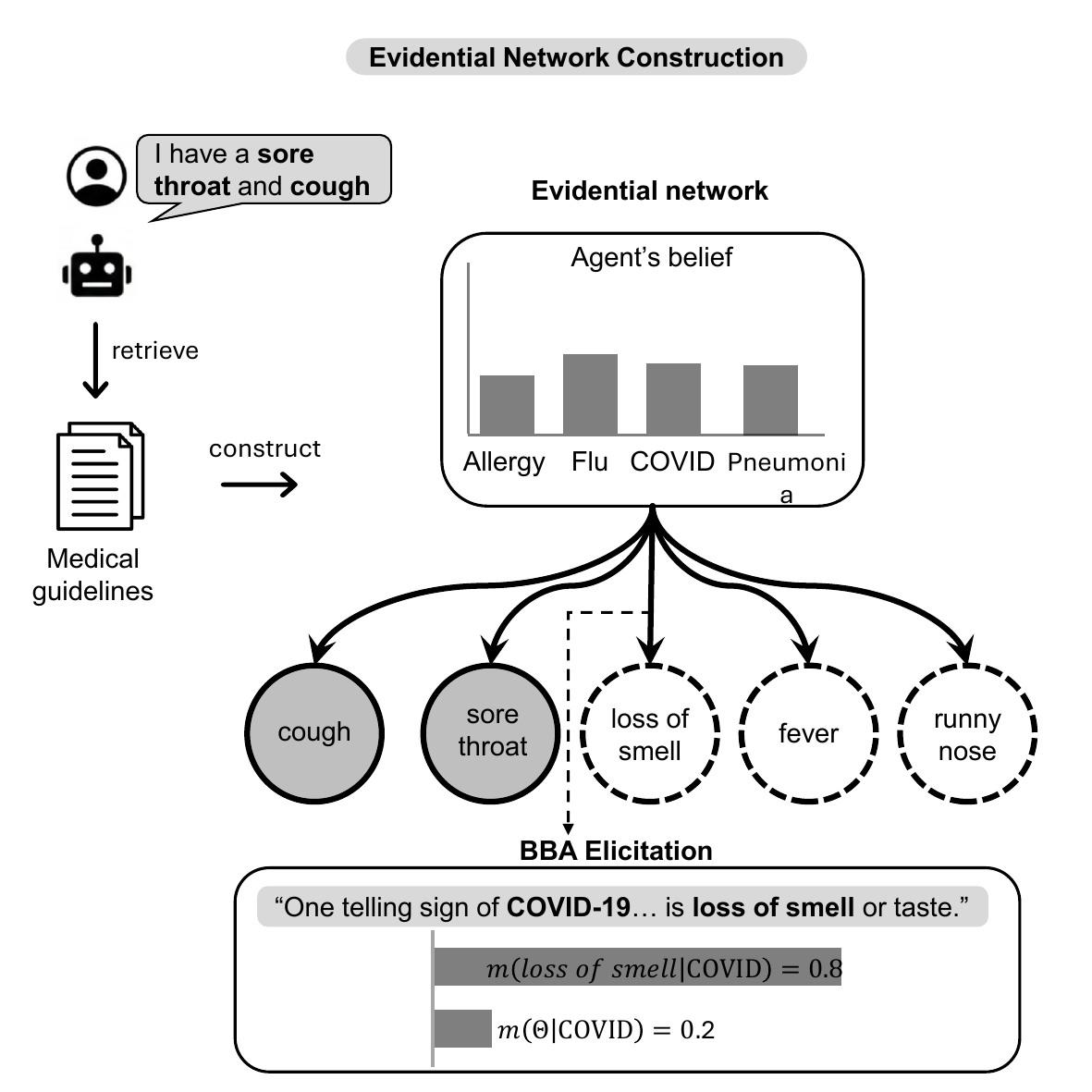}
    \hfill
    \textcolor{gray}{\vrule width 1px}
    \includegraphics[width=0.48\textwidth]{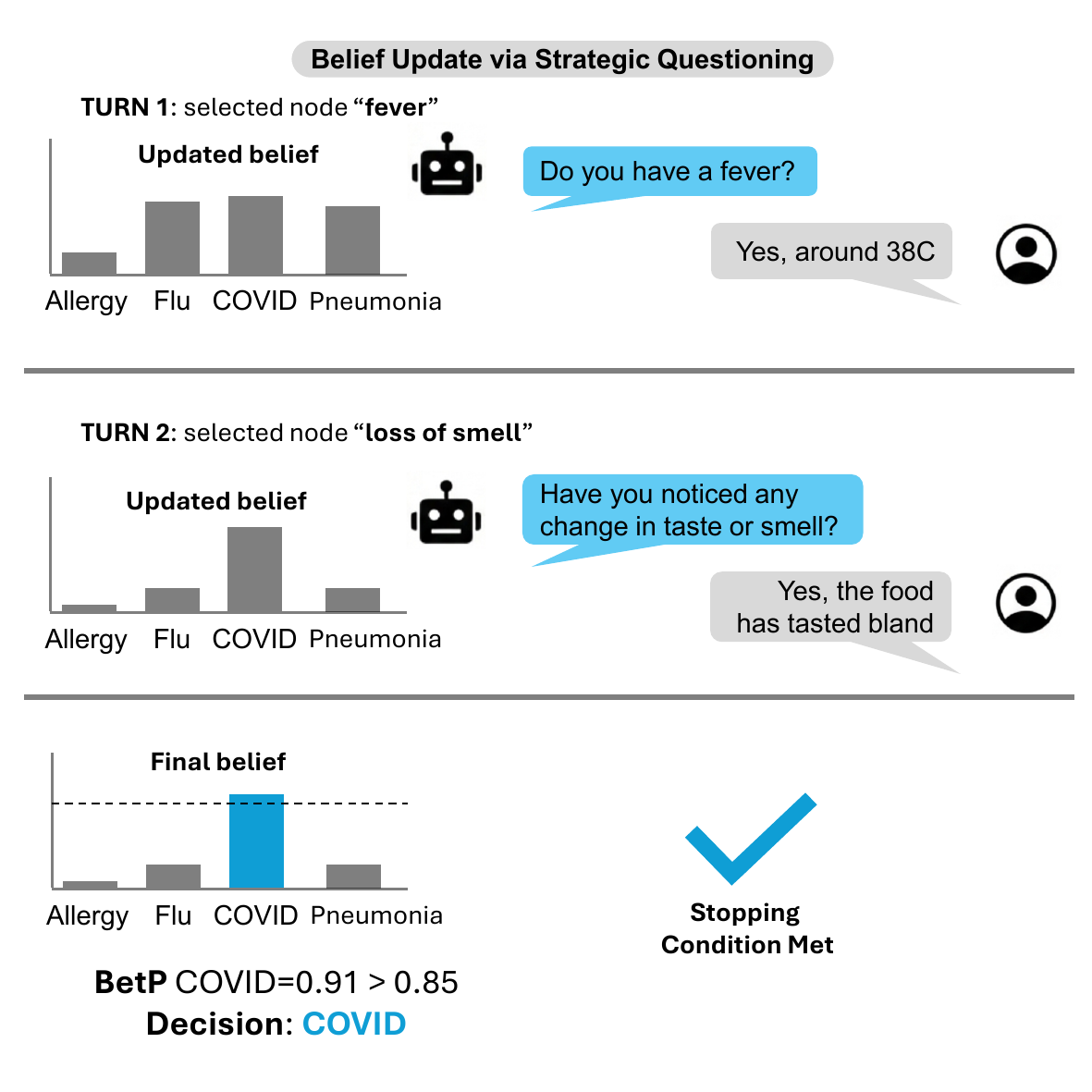}
    \caption{An overview of the \methodname pipeline using a medical diagnosis example. \textbf{Left}: The agent uses the initial query to retrieve documents and construct an evidential network, extracting Basic Belief Assignments (BBAs) from text. \textbf{Right}: The interaction loop where the agent selects an informative node (e.g., 'fever') to generate a targeted question, and updates its belief distribution based on the user's answer. This cycle repeats until the pignistic probability (BetP) of a leading hypothesis meets the stopping condition.}
    \label{fig:pipeline}
\end{figure*}

To guide the generation of question $q_t$, an agent must reason over the joint effect of $\mathcal{D}$ and the anticipated user response on belief updates, addressing ambiguity (e.g., evidence supporting multiple hypotheses), uncertainty, contradictory information, and combinatorial complexity. To address these challenges, we present \methodname, a novel framework for SIA that integrates evidential reasoning with LLMs to systematically reduce uncertainty over a hypothesis space through targeted questioning. As shown in the left panel ('Evidential Network Construction') in Figure~\ref{fig:pipeline}, our approach constructs an evidential network to explicitly model the agent's uncertainty over hypotheses (\S\ref{sec:uncertainty}). Then, during the interaction phase ('Belief Update via Strategic Questioning'), new information from the user is used to update the belief state (\S\ref{sec:method:belief_update}). To guide the questioning strategy, \methodname uses a multi-objective policy (\S\ref{sec:method:question_selection}).

\subsection{Modeling Uncertainty over Hypotheses}
\label{sec:uncertainty}

SIA requires not only representing what an agent knows and does not know, but also reasoning over uncertain and potentially conflicting evidence. Dempster-Shafer (DS) theory and the evidential networks framework provide a principled way to represent and reason in this setting. Unlike Bayesian probability theory, which represents beliefs as a probability distribution over individual hypotheses, DS models belief to be assigned to proper subsets of hypotheses, enabling explicit representation of ambiguity \cite{dempster1967,shafer1976}. DS graphical formalism for belief functions is evidential networks, which extend this capability by offering a mechanism for (1) organizing evidence into a graph of domain variables, such as symptoms and diseases  \cite{xu1994evidential,yaghlane2003directed}; and (2) fusing evidence from different sources while explicitly tracking uncertainty and conflict. Below, we describe how we transform textual evidence from $\mathcal{D}$ into a structured evidential network for downstream reasoning and belief updates.

\paragraph{Network Structure Construction.} We represent domain knowledge with a directed acyclic graph $G=(V,E)$. Each node $Z\in V$ denotes a discrete domain variable with a finite \emph{frame of discernment} $\Theta_Z$ (the set of mutually exclusive states for $Z$). The root $H$ is a decision node, with $\Theta_H \equiv \mathcal{H}$. Other nodes encode intermediate factors (e.g., latent clinical or legal conditions) and observable features (e.g., symptoms, test results, case facts) that can be elicited from the user. The graph is layered, with $H$ at the top, intermediate variables mediating dependencies, and observable variables appearing toward the leaves. We restrict $G$ to be acyclic and of bounded depth to support tractable inference and interpretable structures.

To construct $G$ from $\mathcal{D}$, we perform retrieval-guided expansion starting from $H$. For each hypothesis $h\in\mathcal{H}$, we retrieve passages from $\mathcal{D}$ that are semantically relevant to $h$ and prompt an LLM to propose candidate child variables (e.g., consequences, correlates, refinements) grounded in the retrieved text. Newly added variables are iteratively expanded in the same manner until a depth limit is reached or no further grounded variables are proposed. Appendix~\ref{appx:network_construction} provides the full procedure. The output is a document-grounded graph $G=(V,E)$ with explicit variable definitions and dependencies. In the next subsection, we derive uncertainty-aware edge parameters from the same evidence, yielding a fully parameterized evidential network.

\paragraph{Document-based Belief Elicitation.} The structure $G$ specifies the dependency relations among variables, but it must be complemented with uncertainty-aware parameters that quantify how evidence supports different states. Because textual evidence is often hedged (e.g., ``may indicate''), incomplete, or contradictory, point-valued conditional probabilities can force unjustified commitments. We therefore avoid such representations and instead model evidence using Dempster–Shafer \emph{basic belief assignments} (BBAs). For a node $Z$ with frame $\Theta_Z$, a BBA is a function $m_Z: 2^{\Theta_Z} \rightarrow [0,1]$ that satisfies $\sum_{A \subseteq \Theta_Z} m_Z(A) = 1$ and $m_Z(\varnothing) = 0$. Subsets $A$ with $m_Z(A) > 0$ are called \emph{focal sets}. Mass assigned to singletons $\{z\}$ represents specific support, while mass assigned to proper subsets, including $\Theta_Z$, captures ambiguity or residual ignorance. 

Edges are parameterized using conditional BBAs elicited from evidence in $\mathcal{D}$. For clarity, we consider a single-parent edge $X \to Y$, for each parent state $x \in \Theta_X$ we define a conditional BBA $m_{Y \mid X}(\cdot \mid x): 2^{\Theta_Y} \rightarrow [0,1]$ satisfying $\sum_{A \subseteq \Theta_Y} m_{Y \mid X}(A \mid x) = 1$ and $m_{Y \mid X}(\varnothing \mid x) = 0$. This function specifies how belief over $Y$ is induced when $X = x$. For each edge $X \to Y$, we collect the evidence snippets from $\mathcal{D}$ that motivated introducing this dependency during graph construction and additional retrieved passages that mention both $X$ and $Y$. Given a snippet and a parent state $x$, we prompt the LLM to identify the subset $A \subseteq \Theta_Y$ supported by the snippet under the assumption $X = x$ and to allocate belief mass over focal sets. To keep elicitation tractable while preserving uncertainty, we permit only three forms of focal sets: singletons $\{y\}$ when the evidence is specific, subsets $A$ with $|A| \le 2$ when the snippet supports a limited disjunction, and the full frame $\Theta_Y$. We allocate an explicit ignorance mass to $\Theta_Y$ when the evidence is vague, non-committal, or internally contradictory. Any remaining mass needed to satisfy $\sum_{A \subseteq \Theta_Y} m_{Y\mid X}(A\mid x)=1$ is also assigned to $\Theta_Y$, which both normalizes the BBA and captures residual uncertainty.

Multiple snippets may provide evidence about the same edge $X\rightarrow Y$ under the same conditioning state $x$, producing multiple candidate conditional BBAs $\{m^{(k)}_{Y\mid X}(\cdot\mid x)\}_{k=1}^K$. We aggregate them using Yager's rule of combination \cite{Yager1987}, which preserves contradictions as increased ignorance rather than forcing normalization. For two BBAs $m_1,m_2$ over $\Theta_Y$, we define the conjunctive product:
$$
  m^\cap(A)=\sum_{B\cap C=A} m_1(B)m_2(C), \qquad \forall A\subseteq \Theta_Y,
$$
and let $K = m^\cap(\varnothing)$ be the conflict mass. Yager's combination $m = m_1 \oplus_Y m_2$ is
$$
  m(A)=
  \begin{cases}
    0,                  & A=\varnothing,                         \\
    m^\cap(A),          & A\subset \Theta_Y,\ A\neq \varnothing, \\
    m^\cap(\Theta_Y)+K, & A=\Theta_Y.
  \end{cases}
$$
Yager’s rule is commutative but not associative, therefore, for $K>2$ snippets, we use the standard $K$-way conjunctive combination followed by a single transfer of the resulting conflict mass to $\Theta_Y$. The resulting conditional BBAs $\{m_{Y\mid X}(\cdot\mid x)\}$ form the \static{} evidential parameters used for downstream inference and belief updating.

\paragraph{Validation on Synthetic Ground-Truth Graphs.}
To validate that our pipeline can recover meaningful structure from text, we conduct a controlled benchmark using synthetic text snippets from the QUITE dataset \cite{schrader-etal-2024-quite} generated based on the \texttt{ASIA} Bayesian network \cite{lauritzen1988local}. Each snippet was perturbed with lexical variation and irrelevant distractors. We then run our pipeline with \texttt{gpt-5-nano} to generate evidential networks from scratch. We compare the recovered graphs to the known ground-truth Bayesian structure using Structural Hamming Distance (SHD), Precision, and Recall on the set of directed edges. Across runs, we obtain a mean SHD of 1.4$\pm$0.9 edges, with edge precision 0.91$\pm$0.06 and recall 0.88$\pm$0.08. In some runs, the recovered graph exactly matches the ground truth, and in other runs it is within SHD $\le 2$. Overall, the method consistently recovers plausible graphs with high edge-level accuracy.

\subsection{Evidence Fusion and Belief Updating}
\label{sec:method:belief_update}

During interaction, the agent fuses user answers with the corpus-parameterized evidential network to refine beliefs over $\mathcal{H}$. We represent the agent's belief state at turn $t$ by the marginal BBA at the hypothesis node, $b_t \equiv m_t(H)$, over $G$ given all evidence collected so far (Appendix~\ref{appx:dst}). We obtain the marginal BBA via Shenoy--Shafer local computation, which is the DST analogue of belief propagation (message passing) used in Bayesian networks \cite{shenoy1990axiomatic,xu1994evidential}.

At turn $t$, the agent asks a question $q_t$ targeting a node $Z_t$ and receives an answer $a_t$. We map $a_t$ to an answer BBA $m^{(a_t)}_{Z_t}$ over $\Theta_{Z_t}$ using an LLM constrained to the same focal-set family. Fully specified answers are encoded as singletons (e.g., $m^{(a_t)}_{Z_t}(\{z\})=1$). When the answer expresses uncertainty or hedging, we allocate partial mass to the supported singleton or small subsets and assign the remaining mass to $\Theta_{Z_t}$ to represent residual ignorance. We then fuse this answer evidence with the current local evidence at $Z_t$ using Yager's rule:
$$
  m^{\text{loc}}_{t+1}(Z_t)\;=\;m^{\text{loc}}_{t}(Z_t)\;\oplus_Y\;m^{(a_t)}_{Z_t}.
$$
This fusion routes contradictions to $\Theta_{Z_t}$, preventing overconfident updates under conflicting information. Finally, we rerun belief propagation on $G$ to obtain updated marginals $\{m_{t+1}(Z)\}_{Z\in V}$ and in particular $b_{t+1}=m_{t+1}(H)$.

\subsection{Strategic Question Generation}
\label{sec:method:question_selection}

At each step, the agent selects targeted follow-up questions to reduce uncertainty about $H$. We score candidate questions by the expected decrease in \emph{Deng entropy} \cite{deng2016deng} of $m_t(H)$ after the answer. It decomposes uncertainty into \emph{nonspecificity} (mass on large sets, reflecting ambiguity) and \emph{discord} (competition among focal sets, reducing to Shannon entropy when only singletons are present). Let $\Theta_H$ be the frame of $H$. Deng entropy is:
\begin{align*}
  E_d(m_H) &= \underbrace{\sum_{A\subseteq \Theta} m_H(A)\,\log\!\big(2^{|A|}-1\big)}_{\text{nonspecificity}} \\
  &- \underbrace{\sum_{A\subseteq \Theta} m_H(A)\,\log m_H(A)}_{\text{discord}}
\end{align*}
Our question policy follows a two-stage objective: it first prioritizes reducing nonspecificity (i.e., breaking down broad, ambiguous sets into more specific alternatives), and once nonspecificity is sufficiently small, it prioritizes reducing discord to discriminate among remaining competing hypotheses.

Finally, we employ a process analogous to information gain in Bayesian active learning. Concretely, we consider candidate questions corresponding to nodes $Z$ for which we have not yet elicited an answer. For each candidate $Z$ and each possible answer $z\in\Theta_Z$, we form a deterministic hypothetical answer BBA $m^{(z)}_Z(\{z\})=1$, fuse it into the current local evidence at $Z$, propagate through the network to obtain a hypothetical posterior marginal $m^{(Z=z)}_{t+1}(H)$, and compute the resulting nonspecificity and discord. We weight these outcomes by a predictive distribution over answers, approximated by the current pignistic distribution at $Z$. This yields expected reductions $\Delta_{\text{nonsp}}(Z)$ and $\Delta_{\text{disc}}(Z)$, and we select $Z_t$ by lexicographic maximization of $(\Delta_{\text{nonsp}},\Delta_{\text{disc}})$ with the first component prioritized until nonspecificity at $H$ drops below a preset threshold. The natural-language question $q_t$ is then generated by conditioning an LLM on the selected variable definition and its state descriptions.

Stopping is triggered when the pignistic distribution over hypotheses becomes sufficiently decisive. Given a marginal BBA $m_t(H)$, the pignistic probability of $h\in\Theta_H$ is
$$
  \mathrm{BetP}_t(h)
  =
  \sum_{\substack{A\subseteq \Theta_H\\ h\in A}}
  \frac{m_t(H)(A)}{|A|(1-m(\varnothing))}.
$$
We terminate and output the current best hypothesis when
$$
  \max_{h\in\Theta_H}\mathrm{BetP}_t(h)\ge \tau_{\text{conf}},
$$
for a predefined confidence threshold $\tau_{\text{conf}}$. If this condition is not met by the maximum number of turns $T_{\max}$, the dialogue ends, and the agent abstains.

\section{Experimental Setup}
\label{sec:exp_setup}
\begin{table*}[t]
  \centering
  \small
  \begin{tabular}{@{} l C{1.0cm} C{2.0cm} m{10cm}@{}}
    \toprule
    Domain  & Size & KB & Example \\
    \midrule
    \DomainCell{\MedicalIcon}{Medical} & 1,273 & Medical textbooks & 
    \textbf{Input:} ``While in the ICU, a 62-year-old male undergoes placement of a Swan-Ganz catheter to evaluate his right heart pressures. What are normal values for the pressures that will be obtained from this patient's right ventricle?''\par
    \textbf{Facts:} 1. All pressures are found to be within normal limits.
    2. The cardiology fellow records a pulmonary wedge pressure of 10 mmHg.\par
    \textbf{True Hypothesis:} \texttt{25/5 mmHg} \\
    \midrule
    \DomainCell{\LegalIcon}{Legal} & 211 & legal encyclopedia, secondary sources & 
    \textbf{Input:} ``A homeowner seeks compensation for damage caused by the neighbour's trees. What kinds of orders may the court make?'' \par
    \textbf{Facts:} 1. The properties are adjoining residential lots.
    2. An arborist report attributes cracking to the driveway and damage to stormwater pipes to roots from the neighbour’s eucalyptus trees;\par
    \textbf{True Hypothesis:} \texttt{Trees Act 2006 (NSW) section 9}\\
    \bottomrule
  \end{tabular}
  \caption{Domain-specific statistics and examples in \methodname.}
  \label{tab:domain_stats}
\end{table*}

\subsection{Datasets}
\label{sec:exp_setup:datasets}

Each instance in our datasets consists of 4 components: (1) \textit{initial user query} -- an incomplete and vague description of the user's situation which can serve as a starting point for the conversation; (2) \textit{atomic facts} -- a list of elementary facts about the user’s situation required to identify the correct hypothesis; (3) \textit{relevant documents} -- a set of gold documents containing relevant domain knowledge; and (4) \textit{true hypotheses} -- a list of all hypotheses that the agent needs to identify. \methodname{} has access to a full set of contextual documents and needs to retrieve a subset of \textit{relevant documents}, whereas the \textit{atomic facts} are known only to the client and used to answer the follow-up questions. The \textit{relevant documents} and \textit{true hypotheses} are hidden from both agents and serve only as gold labels for performance evaluation. We test \methodname{} on \textcolor{medColor}{medical} and  \textcolor{lawColor}{legal} datasets described below, with Table~\ref{tab:domain_stats} summarizing their key statistics and example instances.

\paragraph{\MedicalIcon\ Medical Domain.} We use the English portion of MedQA \cite{jin2021disease}, a multiple-choice QA dataset collected from professional medical board exams. We use the correct answer as the true hypothesis (for example, the correct diagnosis). We use the accompanying medical textbooks as a collection of contextual documents. The agent's task is to identify a single true hypothesis by asking targeted questions grounded in both the user's initial input and the evidence found in these medical textbooks.

\paragraph{\LegalIcon\ Legal Domain.} Similarly, we use BarExamQA \cite{zheng2025}, a multiple-choice benchmark built from Multistate Bar Examination (MBE) style questions. Each instance presents a legal fact pattern and a targeted question with four answer choices; we treat these answer choices as the hypothesis set and the correct option as the true hypothesis. BarExamQA additionally provides gold supporting passages and an associated retrieval corpus of paragraph-level legal materials (e.g., case law and secondary sources), which we use as the collection of contextual documents. As in the medical setting, we preprocess BarExamQA to fit our SIA format by constructing an underspecified initial query and a set of atomic facts that are revealed only through follow-up questioning.

\subsection{Dialogue Simulation}
\label{sec:exp_setup:sim}

Similar to prior work \cite{li2024mediq}, \methodname{} represents the domain expert agent (lawyer or doctor), who attempts to identify the \textit{true hypotheses} from the \textit{initial user query} by asking targeted questions to gather missing information, guided by the \textit{contextualized documents} and conversation history. To evaluate \methodname{}, we simulate the client using an LLM that is given access to the \textit{initial user query} and the \textit{atomic facts} and tasked with answering questions from the agent. We include all the prompts used for dialogue simulation in Appendix~\ref{prompt:dialogue-simulation}.

\begin{table*}[!ht]
  \centering
  \small
  \sisetup{
    reset-text-series = false,
    text-series-to-math = true
  }
  \begin{tabular}{
    l
    l
    S S
    c
    S S
  }
    \toprule
    \multirow{2}{*}[-0.5ex]{\textbf{Model}}
      & \multirow{2}{*}[-0.5ex]{\textbf{Method}}
      & \multicolumn{2}{c}{\LegalIcon\normalsize\quad\textbf{Legal domain}}
      & \multicolumn{1}{c}{} 
      & \multicolumn{2}{c}{\MedicalIcon\normalsize\quad\textbf{Medical domain}} \\
    \cmidrule(lr){3-4}\cmidrule(lr){6-7}
      & & {Success (\%) $\uparrow$} & {Avg. Turns $\downarrow$} 
      & & {Success (\%) $\uparrow$} & {Avg. Turns $\downarrow$} \\
    \midrule
    \multirow{5}{*}{\texttt{gpt-5-nano}}
      & AoP   & 17.1 &  2.6 & & 41.6 & 2.1 \\
      & MediQ & 27.2 &  9.4 & & 59.1 & 10.7 \\
      & UoT   & 32.8 & \textbf{4.9} & & 63.8 & 7.6 \\
      & IG Bayesian & 61.3 & 6.2 & & 67.5 & 8.3 \\
      \cmidrule(lr){2-7}
      & \methodname{}
                &   \bfseries 66.5 & 5.9
                & & \bfseries 69.3 & \bfseries 7.4 \\
    \addlinespace[0.2em]
    \midrule
    \multirow{5}{*}{\texttt{Qwen3 32B}}
      & AoP   & 15.4 &  2.2 & & 39.2 & 1.8 \\
      & MediQ & 25.7 &  8.9 & & 56.4 & 9.9 \\
      & UoT   & 60.8 &  5.1 & & 59.5 & 8.3 \\
      & IG Bayesian & 57.9 &  5.7 & & 64.2 & 8.0 \\
      \cmidrule(lr){2-7}
      & \methodname{}
                &   \bfseries 64.3 & 5.4
                & & \bfseries 68.1 & 7.1 \\
  \bottomrule
  \end{tabular}
  \caption{Results of the methods on two datasets across two LLMs.}
  \label{tab:results}
\end{table*}


\subsection{Baselines}
\label{sec:exp_setup:baselines}

We compare \methodname{} against four interactive expert baselines that operate under the same dialogue simulation protocol (\S\ref{sec:exp_setup:sim}) and turn budget $T_{\max}$. All baselines observe the initial user query, the answer options (hypothesis set $\mathcal{H}$), and the dialogue history, and may either ask one follow-up question per turn or produce a final answer. We provide the exact prompts and hyperparameter settings for all baselines in Appendix~\ref{prompt:zero-shot}.

\paragraph{Ask-or-Predict (AoP).} A simple zero-shot baseline that uses an LLM to decide at each turn whether to ask a follow-up question or make a final prediction based on the current information. The model is prompted to generate a single follow-up question if it decides to ask, it directly predicts the final answer.

\paragraph{MediQ \cite{li2024mediq}.} A MediQ-style expert that alternates between (i) an abstention module that estimates whether sufficient information has been gathered to answer confidently and (ii) an atomic question generation module when additional information is needed. The agent explicitly elicits a confidence signal via self-consistency on a verbalized confidence in the final hypothesis to decide whether to ask another question or to answer.

\paragraph{Uncertainty of Thoughts \cite[UoT;][]{hu2024uncertainty}.} A prompt-based method that asks follow-up questions via uncertainty-aware planning. At each turn, the method generates candidate questions, simulates short-term future trajectories, and selects the question that maximizes the expected reduction in uncertainty over the hypothesis set, using an information-gain-motivated reward and reward propagation.

\paragraph{IG Bayesian.} A variant that replaces BBAs with point probabilities extracted from documents. When a probability cannot be derived for a required factor, it defaults to a uniform distribution. This reduces the inference stage to structured Bayesian updating, while retaining the same document usage and expected-information follow-up question selection.

\subsection{Implementation Details}
\label{sec:exp_setup:implementation}

We evaluated \methodname{} using two LLMs: \texttt{gpt-5-nano} \cite{openai2025gpt41}, a popular proprietary model, and \texttt{Qwen3 32B} \cite{qwen3report}, a strong open-source model. For all models, we use a temperature of $0.5$ and nucleus sampling with top-$p$ of 1.0 to balance stability and diversity. We set the turn budget $T_{\max}$ to 15 and the confidence threshold $\tau$ to 0.85 based on our empirical observations.

\subsection{Evaluation Metrics}
\label{sec:exp_setup:metrics}

Our method is evaluated with two complementary metrics. \emph{Success rate} measures the proportion of dialogues where the system identifies the correct true hypothesis. While the metric captures end-task correctness, it can be misleading on its own, as an agent could succeed nearly 100\% of the time simply by asking an unlimited number of follow-up questions. Therefore, we also report \emph{mean dialogue length}, which measures the average number of turns per dialogue.

\section{Results}
\label{sec:results}
Table~\ref{tab:results} reports the success rate and average dialogue length on both legal and medical domains for two backbone LLMs. Overall, \methodname{} achieves a higher success rate while maintaining efficient dialogues across domains and backbone LLMs, consistent with our goal of using explicit, document-grounded uncertainty to inform decision-making.

In the legal domain, \methodname{} with \texttt{gpt-5-nano} attains 66.5\% success. This is a large improvement over AoP (17.1\%) and MediQ (27.2\%), corresponding to gains of 49.4 and 39.3 points. \methodname{} also outperforms UoT by 33.7 points (66.5\% vs. 32.8\%). UoT uses fewer turns (4.9 vs. 5.9), but its substantially lower success rate suggests that optimizing primarily for short dialogues can leave decisive facts unelicited. With \texttt{Qwen3 32B}, we observe the same pattern. \methodname{} achieves 64.3\% success in 5.4 turns, remaining more accurate than UoT (60.8\%) at a comparable dialogue length (5.4 vs. 5.1). In the medical domain, \methodname{} again improves both correctness and efficiency relative to the interactive baselines. With \texttt{gpt-5-nano}, \methodname{} reaches 69.3\% success and the shortest dialogues (7.4 turns), outperforming AoP (41.6\%) and MediQ (59.1\%). It also improves over UoT (63.8\%) while using slightly fewer turns on average (7.4 vs.\ 7.6). With \texttt{Qwen3 32B}, \methodname{} achieves 68.1\% success with 7.1 turns, which is a substantial gain over AoP (39.2\%) and MediQ (56.4\%).

Across LLMs, the success rate is higher in the medical domain than in the legal domain. We hypothesize several factors contributing to this disparity. Medical reasoning relies on biological principles that are globally consistent and supported by vast open-access literature in LLM's pre-training data, whereas legal reasoning is fragmented by jurisdictional variance and proprietary data barriers. Additionally, medical diagnosis often involves clearer, convergent reasoning, unlike the interpretive ambiguity inherent in legal argumentation.

\section{Analysis}
\label{sec:analysis} 
We assess the contribution of components of \methodname{}.

\paragraph{Impact of BBA elicitation.} A key design choice in \methodname{} is to elicit BBAs rather than point-valued probabilities from the retrieved text. This matters because many evidence snippets in both domains are incomplete, hedged, or condition-dependent. In these cases, forcing evidence into a single distribution can introduce artificial certainty and can make belief updates brittle. To assess the contribution of the BBA, we present ablation tests. IG Bayesian in Table~\ref{tab:results} is a variant of \methodname{} that extracts probabilities rather than BBAs, keeping the rest of the pipeline unchanged. When point probabilities cannot be extracted, it falls back to uniform probabilities. Across both backbones and domains, \methodname{} is consistently more accurate and slightly more turn-efficient than this probability-based alternative. In the legal domain, \methodname{} improves success from 61.3\% to 66.5\% with \texttt{gpt-5-nano} while also reducing turns from 6.2 to 5.9. With \texttt{Qwen3 32B}, the gain is larger (57.9\% to 64.3\%), again with a small turn reduction (5.7 to 5.4). In the medical domain, the same trend holds. With \texttt{gpt-5-nano}, success improves from 67.5\% to 69.3\% and turns drop from 8.3 to 7.4. With \texttt{Qwen3 32B}, success rises from 64.2\% to 68.1\% and turns drop from 8.0 to 7.1. These results support the role of uncertainty-aware evidence extraction. The probability-based variant must commit belief mass across single hypotheses even when the underlying text does not justify that commitment, and it must guess when evidence is missing. Both effects can distort early belief states and can lead the question policy to focus on the wrong uncertainties. In contrast, BBAs allow the elicitation step to place mass on small sets or on the full frame when the snippet only supports a disjunction or is too vague to be specific. This preserves uncertainty rather than hiding it, which, in turn, leads to more reliable belief propagation and steadier progress toward a final decision, especially in the legal setting, where small factual distinctions often determine the correct option.

\begin{figure}[!t]
    \centering
    \includegraphics[width=\linewidth]{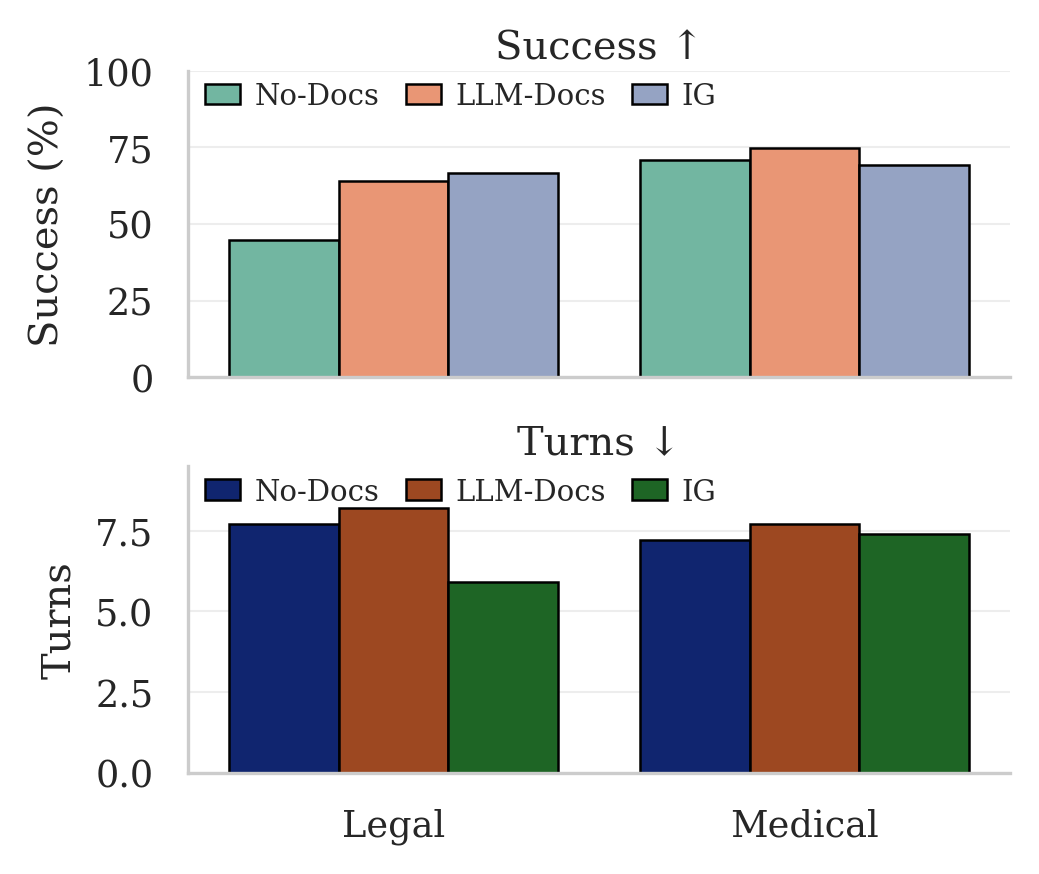}
    \caption{Impact of contextual documents on dialogue performance. With \texttt{gpt-5-nano}, InfoGatherer's (IG) retrieval-grounded evidence improves dialogue efficiency and is especially beneficial in the legal domain, while model-generated references can improve success in the medical domain.}
    \label{fig:docs}
\end{figure}

\paragraph{Impact of contextual documents.} To understand the value of grounding the dialogue in external knowledge, we ablate access to contextual documents. Specifically, we compare \methodname{} to LLM-Docs, which replaces retrieved documents with a model-generated reference document and No-Docs, which removes documents entirely and prompts the LLM to output BBAs directly using only the initial user input and dialogue history. This variant is reported for \texttt{gpt-5-nano} on Fig~\ref{fig:docs}. In the legal domain, \methodname{} is both more accurate and efficient (66.5\% vs. 64.1\%, and 5.9 vs. 8.2 turns). In the medical domain, LLM-Docs achieves higher success (74.6\% vs. 69.3\%) but with similar dialogue length (7.7 vs. 7.4 turns). Overall, these results show that replacing external evidence with model-generated references can yield domain-dependent behavior, while retrieval-grounded evidence provides strong and efficient performance, particularly in the legal setting.

\begin{figure}[t]
    \centering
    \includegraphics[width=\linewidth]{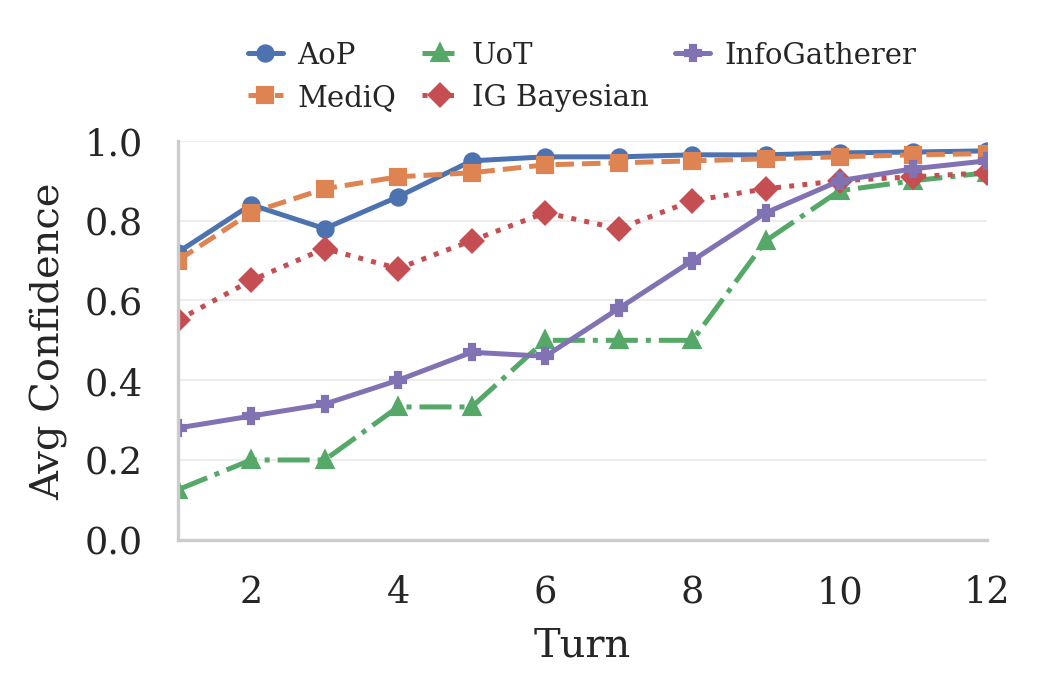}\vspace{-3mm}
    \caption{Average confidence in the correct hypothesis across dialogue turns for five methods, using each method’s native confidence metric. The InfoGatherer's objective, which explicitly targets uncertainty reduction, yields smoother, more information-aligned increases.}
    \label{fig:objectives}
\end{figure}

\paragraph{Dynamics of questioning objective.} We next examine how the choice of questioning objective affects dialogue. Fig.~\ref{fig:objectives} plots each method’s average confidence in the correct hypothesis over turns, using a method-specific confidence metric, so we compare trajectories rather than absolute calibration. Overall, AoP, MediQ, and IG Bayesian exhibit varying degrees of overconfidence. AoP, which lacks an explicit value-of-information criterion, exhibits noisy trajectories with abrupt jumps and early plateaus, reflecting myopic, confirmation-biased questioning and premature commitment once a plausible narrative emerges. MediQ ties stopping to internal self-consistency rather than to external discriminative evidence, producing smooth, steadily increasing confidence that can be misleading, as agreement across samples may rise through rationalization rather than the acquisition of new information. UoT’s subset elimination-based objective yields a characteristic staircase pattern because confidence is proportional to $1/|H_t|$ when the true hypothesis remains feasible, progress occurs only when questions eliminate portions of the hypothesis set, leading to extended plateaus when remaining hypotheses are difficult to separate. IG Bayesian replaces structured uncertainty with point probabilities, yielding smoother updates but increased brittleness, as miscalibrated probability extractions can induce sharp swings, whereas a uniform fallback for missing values flattens posteriors and weakens question selection under ambiguity. In contrast, InfoGatherer aligns its questioning objective with the structure of uncertainty itself, first prioritizing reductions in nonspecificity by moving belief mass off composite sets, and then resolving discord by targeting conflicts among specific hypotheses. This strategy accounts for the conservative early phase, followed by sustained late-turn gains in confidence, yielding calibrated progress without early saturation. Overall, the comparison shows that objectives based on internal agreement or uniform point probabilities often appear confident without reliably improving correctness, whereas explicitly modeling and reducing different forms of uncertainty supports more informative questions, continued progress, and more reliable stopping decisions.

\section{Related Work}
\label{sec:related_work}
\paragraph{Follow-up question asking.} Earlier work in NLP generated questions answerable by a given text, for reading comprehension and educational setups \cite{du-etal-2017-learning,uto-etal-2023-difficulty,luo-etal-2024-chain}. Another line of work focused on generating follow-up questions that seek additional information not previously answered \cite{majumder-etal-2021-ask,meng-etal-2023-followupqg,liu2025bridging}. With the advent of LLMs and their use as chatbots across domains, there is growing interest in information-seeking dialogues. In this setup, LLMs may be used to clarify user intent \cite{li2024mediq,wu-etal-2023-inscit,zhang2024modelingfutureconversationturns} or gather user preferences \cite{li2025personalized,bose2026cold,li2023eliciting,andukuri2024stargate}. Despite progress, the ability to deploy these systems in specialized domains is still limited. There is little work on healthcare applications, with most focusing on medical diagnosis using synthetic datasets \cite{wu2023large,li2024mediq,hu2024uncertainty}. Moreover, existing systems still struggle to recognize whether and what information is missing \cite{johri2023testing,tu2024conversational}. 
Meanwhile, multi-agent and human-AI collaboration frameworks \cite{stanford1984diagnostic,wu2023autogen,zhou2024sotopia, wu2024coworkers, deng2024humancentered, lin2024decisionoriented, wang2024healthq, yao2025intelligent} have showcased impressive interactive capabilities and stand to benefit from more robust question-asking techniques. In response, we propose a framework designed to enhance the information-seeking skills of conversational agents across diverse domains. By focusing on systematically eliciting additional details and abstaining when appropriate, our work opens new avenues for advancing question-asking methodologies in high-stakes scenarios and beyond.

\paragraph{Dempster-Shafer Theory in NLP.} Dempster-Shafer Theory has been used in NLP as a way to represent and combine uncertain information when interpreting language. Prior work has used DST in mixed-initiative dialogue to predict conversational initiative based on lexical and discourse cues \cite{chu-carroll-brown-1997-tracking}, in lexical semantics to infer missing semantic relationships \cite{goldhahn-quasthoff-2010-automatic}, in extractive summarization to model uncertainty in sentence salience judgments \cite{manna-etal-2012-subjective}, in authorship attribution to assign belief to subsets of labels \cite{patchala-bhatnagar-2018-authorship}, in socially grounded language understanding to represent norms and interpret indirect speech acts \cite{8268261,Wen_Siddiqui_Williams_2020}, and in stance detection to aggregate rhetorical evidence into interpretable predictions \cite{saha-etal-2024-stance}. Across these applications, basic belief assignments have been estimated from frequencies, features, or discourse cues, and then fused to form robust interpretations. While these studies highlight DST's value for fusing heterogeneous evidence and preserving uncertainty, they rely on heuristic mappings from text to belief. By integrating DST with LLMs, we enable principled uncertainty-aware aggregation and reasoning over text.

\section{Conclusion}
\label{sec:conclusion}
We proposed \methodname{}, a principled framework for conversational information gathering that combines evidence from contextual documents with user inputs to model uncertainty and guide question selection. \methodname{} employs Dempster-Shafer theory together with LLMs to elicit belief mass assignments to guide follow-up question generation. By doing so, our approach enables agents to identify and fill information gaps before making predictions.

Across legal and medical benchmarks, \methodname{} consistently outperforms strong baselines, achieving higher success rates with similar or fewer dialogue turns. Ablation studies demonstrate that each component, document‑grounded belief elicitation, principled questioning, retrieval of external sources, and evidential networks, contributes to these gains. 

This work highlights the importance of integrating formal uncertainty reasoning into LLM-driven dialogue agents, particularly in high‑stakes domains where reliability is paramount. Future work could extend our framework to open‑domain tasks with larger document collections, explore adaptive retrieval strategies, and investigate richer models of user behaviour and preferences.  
By continuing to close the gap between statistical language models and principled decision-making, we hope to enable the safer and more effective deployment of conversational AI systems.

\section*{Limitations}
\label{sec:limitations}
While promising, our study has limitations. The experiments use curated document collections and simulated users, which may not capture the full complexity of real‑world interactions.  
Extending the approach to open‑domain settings with large and noisy document corpora is a challenging direction. Moreover, our current implementation assumes a fixed set of hypothesis variables; adapting \methodname{} to tasks with dynamic or continuous outcomes remains an open problem. In future work, we aim to explore adaptive retrieval strategies and richer models of user behaviour.

\section*{Ethical Considerations}
\label{sec:ethics}
\paragraph{Data.} The benchmarks used in our work, MedQA and BarExamQA, and their corresponding retrieval corpora, are publicly available. The questions in both benchmarks are sourced from professional exams. 

\paragraph{Responsible Future Deployment.} We tested \methodname{} on QA benchmarks. Care should be taken when deploying systems for real-world usage in the medical and legal domains. Rather than automating expert decision-making, we propose using \methodname{} to support domain experts in gathering the required information for the decision. For example, it can be used to gather information from a patient waiting in the emergency department about their medical concern and produce a summary for the nurse examining the patient. 

While we tested \methodname{} for its ability to predict a label (e.g., a medical diagnosis or a legal ruling), it could also be used to enhance RAG systems that generate answers. \methodname{} is designed to resolve ambiguities and reduce uncertainty, which could lead to improved accuracy in generative question answering. That said, any system that generates text with LLMs can hallucinate. 

\paragraph{Fairness.} Medical patients and legal clients alike tend to receive differing quality of service depending on various aspects of identity, including race, gender, and socioeconomic class \cite{medicalbias1,biasinlaw1}. \methodname{} 
can potentially mitigate these issues by introducing an unbiased interface to medicine and law, but it also risks amplifying bias without careful design. \methodname{} uses LLMs, which are pre-trained on web text that exhibits societal and cultural biases \cite{hershcovich-etal-2022-challenges,liu2025badworktimecrosscultural}. Deploying this system for real-world use would require developing measures to ensure -- and continuously monitor -- that the system has a positive impact and that the outcome of diverse users is equally fair. 

\section*{Acknowledgments}
This work was funded, in part, by the Vector Institute, Canada CIFAR AI Chairs program, NSERC Discovery and Alliance grants, and Accelerate Foundation Models Research Program Award from Microsoft. Additionally, this research was developed in part with funding from the Defense Advanced Research Projects Agency's (DARPA) SciFy program (Agreement No. HR00112520300). The views expressed are those of the author and do not reflect the official policy or position of the Department of Defense or the U.S.~Government. This material is based in part upon work supported by the Defense Advanced Research Projects Agency and the Air Force Research Laboratory, contract number(s): FA8650-23-C-7316. Any opinions, findings and conclusions, or recommendations expressed in this material are those of the author(s) and do not necessarily reflect the views of AFRL or DARPA. This research was supported by the University of Washington Population Health Initiative, Amazon Health, the UW+Amazon Science Hub. 

\bibliography{custom,anthology-1,anthology-2}

\clearpage
\onecolumn
\appendix
\section{Mathematical Framework of Dempster-Shafer Theory}
\label{appx:dst}

Let $G=(V,E)$ be a directed graph. Each node $Z\in V$ has a finite set of mutually exclusive, collectively exhaustive states $\Theta_Z$ (its \emph{frame}). Evidence about $Z$ is represented by a basic belief assignment (BBA) $m_Z:2^{\Theta_Z} \rightarrow[0,1]$ with $\sum_{A\subseteq\Theta_Z} m_Z(A)=1$. To keep induction and downstream inference efficient, we restrict focal sets to singletons, a small number of size-two subsets, and $\Theta_Z$.

Our evidential network is a directed graph $G=(V,E)$, where each node $Z\in V$ is a discrete domain variable with a finite frame of discernment $\Theta_Z$ that lists its mutually exclusive and collectively exhaustive states (e.g. $\Theta_{fever} = \{\texttt{Absent}, \texttt{Mild}, \texttt{Severe}\}$). The root of the network is the decision node $H$ with frame $\Theta_H \equiv \mathcal{H}$, corresponding to the hypothesis space. The network provides reasoning scaffolding where internal variables (e.g. symptoms or legal facts) mediate how evidence relates to $H$. For each node $Z$ we attach a basic belief assignment (BBA) $m_Z:2^{\Theta_Z}\!\to[0,1]$ that distributes belief mass over subsets of the frame $\Theta_Z$ and satisfies $\sum_{A\subseteq \Theta_Z} m_Z(A)=1$. Assigning mass to sets (for example $\{z_i,z_j\}$) encodes ambiguity without forcing premature commitment to a single state. From $m_Z$ we derive standard summary quantities: the belief in $A$ is $\Bel_Z(A)=\sum_{B\subseteq A} m_Z(B)$; the plausibility of $A$ is $\Pl_Z(A)=\sum_{B\cap A\neq\varnothing} m_Z(B)$; and the node's ignorance is $m_Z(\Theta_Z)$, which represents the total unresolved uncertainty at $Z$. In practice, we restrict the focal sets (those with nonzero mass) to singletons, a small number of size-two subsets, and $\Theta_Z$ to keep inference efficient while preserving the ability to represent ambiguity. The node-local BBAs represent evidence extracted from documents and user responses, and their uncertainty.

\begin{tcolorbox}[title={DST imprecision vs.~ignorance vs.~conflict},colback=gray!3,colframe=gray!40]
Let $\Theta$ be the frame of discernment and $m:2^{\Theta}\to[0,1]$ a basic belief assignment (BBA) with
$\sum_{A\subseteq \Theta} m(A)=1$ and $m(\varnothing)=0$.
Mass on different sets has distinct semantics:

\smallskip
\noindent\textbf{Imprecision (partial specificity).}
Evidence supports a \emph{proper} subset $A\subset \Theta$ but does not resolve elements inside $A$.
Example for $\Theta=\{a,b,c\}$: ``could be $a$ or $b$''
$$
m(A)=m(\{a,b\})=0.7,\qquad m(\Theta)=0.3.
$$
Here $m(\{a,b\})$ encodes disjunctive support (imprecision), while $m(\Theta)$ encodes residual ignorance.

\smallskip
\noindent\textbf{Ignorance / non-discriminative evidence.}
Evidence provides no discrimination among hypotheses, so it assigns mass to the whole frame:
$$
m(\Theta)=1.
$$
This is \emph{not} conflict; it is explicit uncertainty about which element of $\Theta$ holds.

\smallskip
\noindent\textbf{Conflict (contradictory evidence).}
When combining BBAs conjunctively, contradiction appears as mass accumulated on the empty set under the
unnormalized conjunctive rule
$$
(m_1 \,\otimes\, m_2)(C)\;=\;\sum_{A\cap B=C} m_1(A)m_2(B),
\qquad K \equiv (m_1 \,\otimes\, m_2)(\varnothing)=\sum_{A\cap B=\varnothing} m_1(A)m_2(B).
$$
Conflict $K$ should be tracked explicitly (or retained in the unnormalized result), and should not be
silently reallocated to $m(\Theta)$. Under normalized Dempster combination,
$m_1\oplus m_2$ rescales non-empty masses by $1/(1-K)$, which is a \emph{normalization} operation rather than
a reinterpretation of conflict as ignorance.
\end{tcolorbox}

Formally the constructed graph $G=(V,E)$ is called a directed evidential network in which each node $Z \in V$ carries a basic belief assignment (BBA) $m_Z$ over its frame of discernment $\Theta_Z$, 
and each edge $X\!\to\!Y \in E$ carries a conditional BBA $m_{Y\mid X}$ that encodes how evidence on $X$ informs beliefs on $Y$ 
\cite{xu1994evidential,yaghlane2003directed}. 
This representation generalizes Bayesian networks by allowing mass on subsets of states, thereby supporting ambiguity and explicit ignorance \cite{shafer1976,smets1993}. 

\paragraph{Inference.} Inference in such networks is performed by local message passing. Incoming evidence at a node is transformed along edges using the conditional rules, either via ballooning extension of the conditionals together with vacuous extension of the parent evidence \cite{smets1993,xu1996reasoning}, or equivalently via the disjunctive rule of combination for parent $\to$ child messages and the generalized Bayesian theorem for child $\to$ parent messages \cite{smets1993}. The resulting messages are fused with the node's current BBA using the conjunctive rule of combination (Dempster's rule in its normalized form \cite{dempster1967,shafer1976}). The evidence conflict $K$ is tracked as a signal of inconsistency between sources \cite{smets1990}. This process yields a marginal BBA at the decision node $H$, corresponding to the hypothesis space. From this marginal we compute the pignistic distribution $\BetP_H(h)=\sum_{A\ni h} m_H(A)/|A|$ for the final decision-making \cite{smets1990}. This distribution provides the feedback signals used by the control policy to guide strategic questioning and to decide when to stop once a sufficient level of confidence has been reached.

\section{Evidential Network Construction}
\label{appx:network_construction}

\begin{figure}[t]
    \centering
    \begin{framed}
    \footnotesize
    \begin{algorithmic}[1]
    \Require corpus $\mathcal{D}$, queue $Q$ (seeded with $H$), graph $G=(V,E)$
    \While{$Q \neq \emptyset$}
      \State $X \gets \Call{Dequeue}{Q}$
      \State $R \gets \Call{Retrieve}{\mathcal{D}, X}$ \Comment{passages}
      \State $\mathcal{C} \gets \Call{GetChildrenLLM}{X, R}$
      \ForAll{$Y$ in $\mathcal{C}$}
        \If{$Y \notin V$ \textbf{ and } not \Call{Cyclic}{$G, X, Y$}}
            \State $V \gets V \cup \{Y\}$ \Comment{add node}
            \State $E \gets E \cup \{(X,Y)\}$ \Comment{add edge}
            \State \Call{Enqueue}{$Q, Y$}
        \EndIf
      \EndFor
    \EndWhile
    \State \Return $G$
    \end{algorithmic}
    \end{framed}
    \vspace{-\baselineskip}
    \captionof{algorithm}{Constructing the evidential network.}
    \label{alg:graph-construction}
    \vspace{-\baselineskip}
\end{figure}

To construct $G$ from the corpus $\mathcal{D} = \{d_1, \dots, d_n\}$, we use a retrieval-guided breadth-first expansion procedure (Algorithm 1). The algorithm initializes a queue $Q$ with the root node $H$ and an empty edge set $E$. At each iteration, it dequeues a variable $X$ from $Q$ and retrieves passages from $\mathcal{D}$ that are semantically related to $X$. These passages are passed to an LLM, which proposes a set of candidate child variables ${Y}$ along with snippets that justify potential dependencies between $X$ and each $Y$. Intuitively, the LLM identifies concepts that appear in the corpus as consequences, correlates, or refinements of $X$.

For each candidate child $Y$, we update the graph as follows. If $Y$ is not already present in $V$ and adding the directed edge $(X, Y)$ does not introduce a cycle, we insert $Y$ into $V$, add $(X, Y)$ to $E$, and enqueue $Y$ into $Q$ for further expansion. The expansion terminates when the queue is empty or a predefined maximum depth is reached. This procedure yields a layered structure rooted at $H$, in which every edge is backed by at least one concrete snippet from $\mathcal{D}$ that motivated its inclusion.

To maintain a compact and meaningful graph, we apply simple regularization and quality-control steps during construction. First, we enforce acyclicity and optionally cap the in-degree and out-degree of nodes to avoid overly dense connectivity. Second, we merge semantically redundant variables (e.g., paraphrases or near-duplicate labels) proposed for different parents into a single node. Third, we can impose type constraints when appropriate (for example, preventing certain observable nodes from becoming parents of the hypothesis node if this conflicts with domain knowledge).

\section{Legal Dataset Preprocessing}
\label{appx:legal_dataset}

In the legal domain, we base our dataset on BarExamQA \cite{zheng2025}, which evaluates legal reasoning and professional judgment using MBE-style multiple-choice questions. We convert each BarExamQA item into an SIA instance following the same preprocessing pattern as MedQA: (i) we derive an intentionally underspecified \textit{initial user input} from the question’s fact pattern; (ii) we extract \textit{atomic facts} capturing the key legally relevant details that may be elicited through dialogue; (iii) we use BarExamQA’s associated legal passage pool as \textit{contextual documents} and treat the annotated supporting passage(s) as gold \textit{relevant documents}; and (iv) we define \textit{hypotheses} as the four answer choices, with the correct choice as the \textit{true hypothesis}. Finally, we use LLM-as-Judge to filter out incorrect and incomplete examples.

\section{Prompts}

We use dspy signatures to represent prompts and parse the model's output.

\label{appx:prompt}

\begin{prompt}[title={Prompt \thetcbcounter: Client Simulation}, label=prompt:dialogue-simulation]
\begin{lstlisting}
class Client(dspy.Signature):
    """You are a truthful assistant that understands client's information. You are given a set of atomic facts about a client and a question from an expert. Select at most two facts that answer the question. If no fact answers the question, return an empty list. Answer only what the question asks for.
    """

    facts: list[str] = dspy.InputField(desc="The client's factual information.")
    question: str = dspy.InputField(desc="The expert's question.")
    response: list[str] = dspy.OutputField(
        desc="At most two facts that answer the expert's question or an empty list if no fact answers the question."
    )
\end{lstlisting}
\end{prompt}

\begin{prompt}[title={Prompt \thetcbcounter: Ask-or-Predict}, label=prompt:zero-shot]
\begin{lstlisting}
class AskOrPredict(dspy.Signature):
    """Answer the inquiry about the case only if all the required information to answer it correctly and factually is explicitly provided in the dialogue history. If the required information is missing, ask a single, atomic information-seeking question to elicit the missing information. Never ask questions that have already been asked in the dialogue history, instead, ask an alternative question. Do not guess or infer the answer from the incomplete information unless there no more questions to ask. Respond only with a single, atomic question or an answer letter if the answer can be determined from the dialogue history."""

    case_inquiry: str = dspy.InputField(desc="The inquiry about the case.")
    dialogue_history: list[dict] | None = dspy.InputField(
        desc="The dialogue history between the expert and the client."
    )
    hypothesis: list[str] = dspy.InputField(desc="The answer options to choose from.")
    response: str = dspy.OutputField(
        desc="A single, atomic question or an answer letter."
    )
\end{lstlisting}
\end{prompt}

\begin{prompt}[title={Prompt \thetcbcounter: BBA elicitation}, label=prompt:mass-fn]
\begin{lstlisting}
class BBAExtractor(dspy.Signature):
    """
    Elicit a conditional belief assignment for an edge X -> Y from a single evidence snippet, under the assumption that X = x.
    """

    snippet: str = dspy.InputField(desc="Evidence snippet mentioning X and Y.")
    parent_var: str = dspy.InputField(desc="Parent X.")
    parent_state: str = dspy.InputField(desc="Assume parent is in x state.")
    child_var: str = dspy.InputField(desc="Child Y.")
    child_frame: List[str] = dspy.InputField(desc="List of all states of the child Y.")
    support: Dict[str, str] = dspy.OutputField(desc="What outcome and how strongly the evidence supports them.")
\end{lstlisting}
\end{prompt}

\begin{prompt}[title={Prompt \thetcbcounter: Fact Generation}, label=prompt:atom-generation]
\begin{lstlisting}
class FactExtractor(dspy.Signature):
    """You are an expert linguist. You are given a context and a sentence from the context. Generate a list of atomic facts that are strictly logically entailed by the given sentence. Keep each fact independent and self-contained. Each fact should make sense when read on its own. Only write facts directly described or supported by the sentence.
    """

    context: str = dspy.InputField()
    sentence: str = dspy.InputField()
    facts: list[str] = dspy.OutputField(desc="A list of atomic facts.")
\end{lstlisting}
\end{prompt}

\section{Baselines Details}
\label{appx:baselines}

\subsection{MediQ}
To evaluate MediQ in the medical domain, we run the code from the author's repository \url{https://github.com/stellalisy/mediQ}. We used the best-reported expert in the paper, which utilizes the scale-based abstention module with rationale generation and a self-consistency factor of 3. We set confidence thresholds to 0.95 that we found achieves the highest success rate on the dev split of iMediQ. For evaluation in the legal domain, we make minor prompt modifications to adapt the medical expert into a legal expert, while leaving the rest of the implementation unchanged.

\subsection{Uncertainty of Thoughts}
We follow the default settings suggested in UoT's paper and the corresponding GitHub repository \url{https://github.com/zhiyuanhubj/UoT}. UoT is a prompt-based uncertainty-aware planning method that generates candidate follow-up questions, simulates short future trajectories for each candidate, and selects the question with the highest expected uncertainty reduction using an information-gain-motivated reward and reward propagation. We set the number of branches for the method \texttt{n\_potential\_actions}=3 and the \texttt{depth}=2. To evaluate UoT in the legal domain, we made minor prompt edits to adapt the medical scenario framing into a legal one, while keeping the UoT algorithm and all other implementation details unchanged.

\end{document}